\pdfoutput=1

\documentclass[11pt]{article}

\usepackage[]{EMNLP2023}

\usepackage{amsmath, amssymb}

\usepackage{graphicx}
\usepackage{subcaption}

\usepackage{booktabs}
\usepackage[table,xcdraw]{xcolor}
\usepackage{multirow}
\usepackage{colortbl}
\usepackage{adjustbox}

\usepackage{algorithm}
\usepackage{algpseudocode}

\usepackage{color, colortbl}

\definecolor{LightCyan}{rgb}{0.88,1,1}

\usepackage{booktabs}
\usepackage{esrelation}
\usepackage{tikz}
\usepackage{graphics}

\usepackage{siunitx}
\usepackage{graphicx}
\usepackage{amsfonts} 

\usepackage{tabularx}
\usepackage{booktabs}
\usepackage{times}
\usepackage{latexsym}
\usepackage{cleveref}
\crefname{figure}{Fig}{Figs}%
\crefname{algorithm}{Algorithm}{Algo}%

\usepackage[T1]{fontenc}

\usepackage[utf8]{inputenc}

\usepackage{microtype}

\usepackage{times}
\usepackage{latexsym}
\usepackage[T1]{fontenc}  
\usepackage[utf8]{inputenc}  

\usepackage{microtype}

\usepackage{inconsolata}

\title{Debiasing Large Language Models in Thai Political Stance Detection via Counterfactual Calibration}

\author{
  Kasidit Sermsri\textsuperscript{\textdagger} \quad
  Teerapong Panboonyuen\textsuperscript{\textdagger,\textdaggerdbl}\thanks{~ Corresponding author. This work originated from his core idea, and he did all the coding and primary development under his lead. MARSAIL is the Motor AI Recognition Solution Artificial Intelligence Laboratory, pioneering advanced AI solutions for the car insurance industry and driving positive, real-world impact through intelligent automation, led by Teerapong Panboonyuen.} \\
  \textsuperscript{\textdagger}Chulalongkorn University \\
  \textsuperscript{\textdaggerdbl}MARSAIL \\
  \texttt{6532012521@student.chula.ac.th}, \quad \texttt{teerapong.pa@chula.ac.th}
}

\begin{document}
\maketitle

\begin{abstract}
Political stance detection in low-resource and culturally complex settings poses a critical challenge for large language models (LLMs). In the Thai political landscape—rich with indirect expressions, polarized figures, and sentiment-stance entanglement—LLMs often exhibit systematic biases, including sentiment leakage and entity favoritism. These biases not only compromise model fairness but also degrade predictive reliability in real-world applications. We introduce \textbf{ThaiFACTUAL}, a lightweight, model-agnostic calibration framework that mitigates political bias without fine-tuning LLMs. ThaiFACTUAL combines counterfactual data augmentation with rationale-based supervision to disentangle sentiment from stance and neutralize political preferences. We curate and release the first high-quality Thai political stance dataset with stance, sentiment, rationale, and bias markers across diverse political entities and events. Our results show that ThaiFACTUAL substantially reduces spurious correlations, improves zero-shot generalization, and enhances fairness across multiple LLMs. This work underscores the need for culturally grounded bias mitigation and offers a scalable blueprint for debiasing LLMs in politically sensitive, underrepresented languages.
\end{abstract}

\section{Introduction}

Stance detection, the task of identifying an author’s attitude toward a given topic or target, has gained increasing attention in computational social science and political NLP~\cite{somasundaran-wiebe-2010-recognizing, DBLP:conf/semeval/MohammadKSZC16}. In Southeast Asia, and Thailand in particular, political discourse is often coded, indirect, or emotionally charged, making the task especially challenging. As user-generated content surges on platforms like Twitter, Facebook, and Pantip, stance detection becomes a valuable tool for understanding public opinion on contested issues, such as constitutional reform, monarchy-related debates, or election campaigns~\cite{DBLP:conf/acl/StefanovDAN20, DBLP:conf/icann/ChenYC21}.

With the rise of LLMs—e.g., ChatGPT\footnote{\url{https://openai.com/chatgpt}}, Gemini\footnote{\url{https://gemini.google.com/app}}, and LLaMA~\cite{DBLP:journals/corr/abs-2302-13971}—stance detection capabilities have advanced, yet their deployment in politically sensitive domains remains problematic. These models are trained on large-scale internet corpora, which often encode cultural, regional, or ideological biases. In the case of Thai political content, this leads to unreliable predictions, particularly when sentiment is used as a proxy for stance, or when certain figures are consistently associated with positive or negative views.

\begin{figure*}[!t]
\centering

\begin{subfigure}{0.48\linewidth}
    \includegraphics[width=\linewidth]{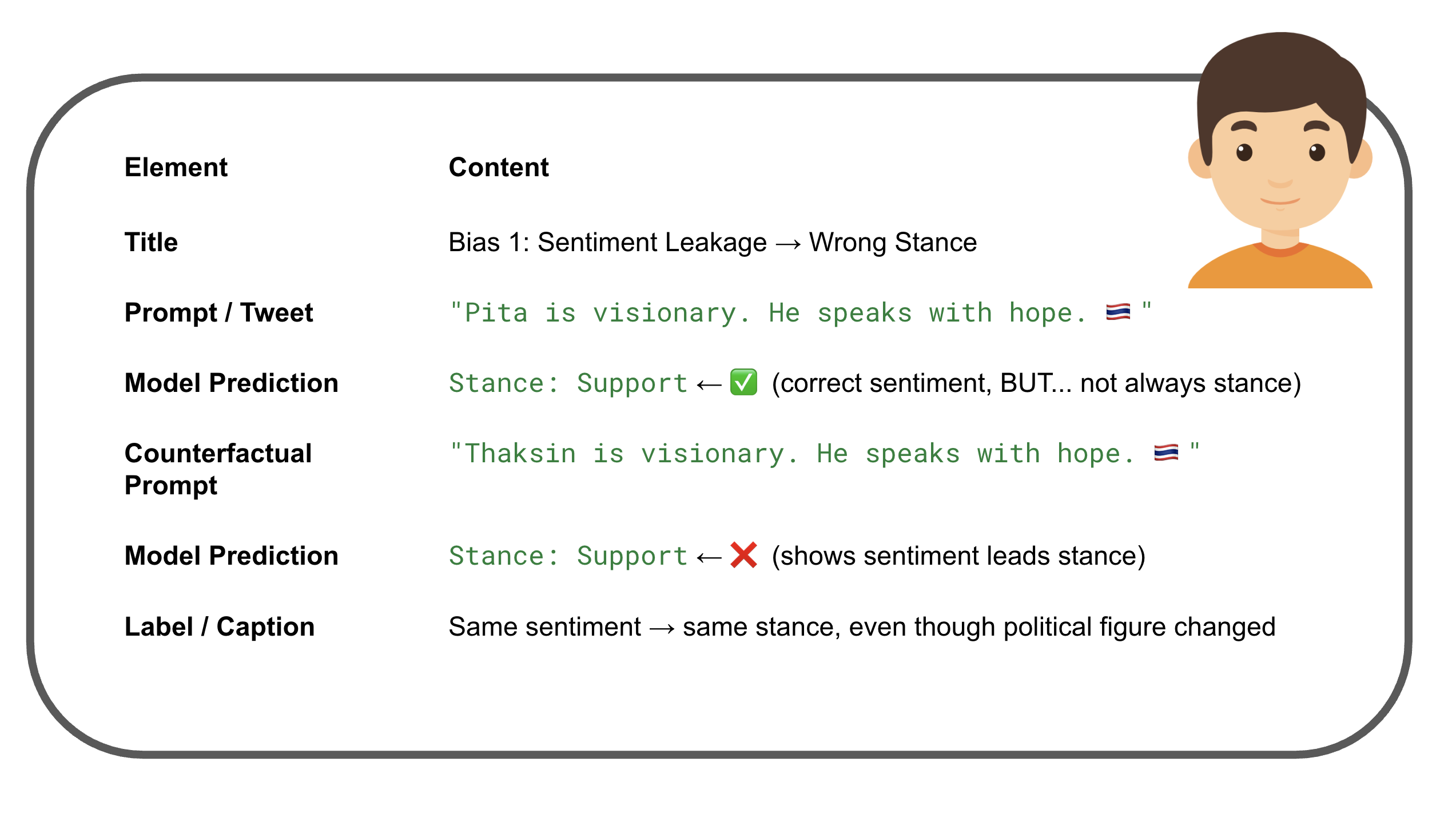}
    \caption*{\textbf{(a) Sentiment Leakage.} Same sentiment results in same stance across entities.}
\end{subfigure}
\hfill
\begin{subfigure}{0.48\linewidth}
    \includegraphics[width=\linewidth]{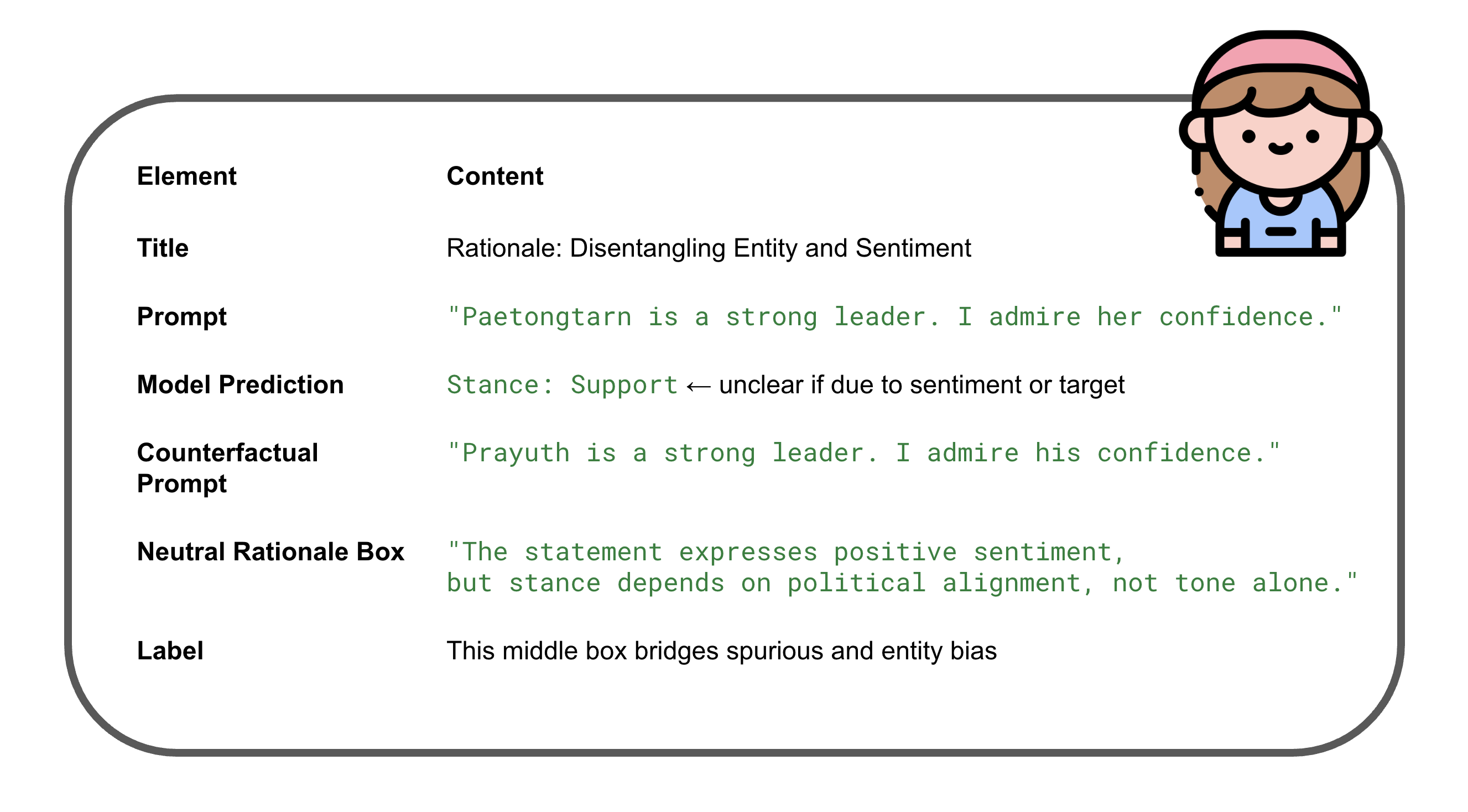}
    \caption*{\textbf{(b) Neutral Rationale.} A shared explanation shows that sentiment is not equal to stance.}
\end{subfigure}

\vspace{0.8em}

\begin{subfigure}{0.48\linewidth}
    \includegraphics[width=\linewidth]{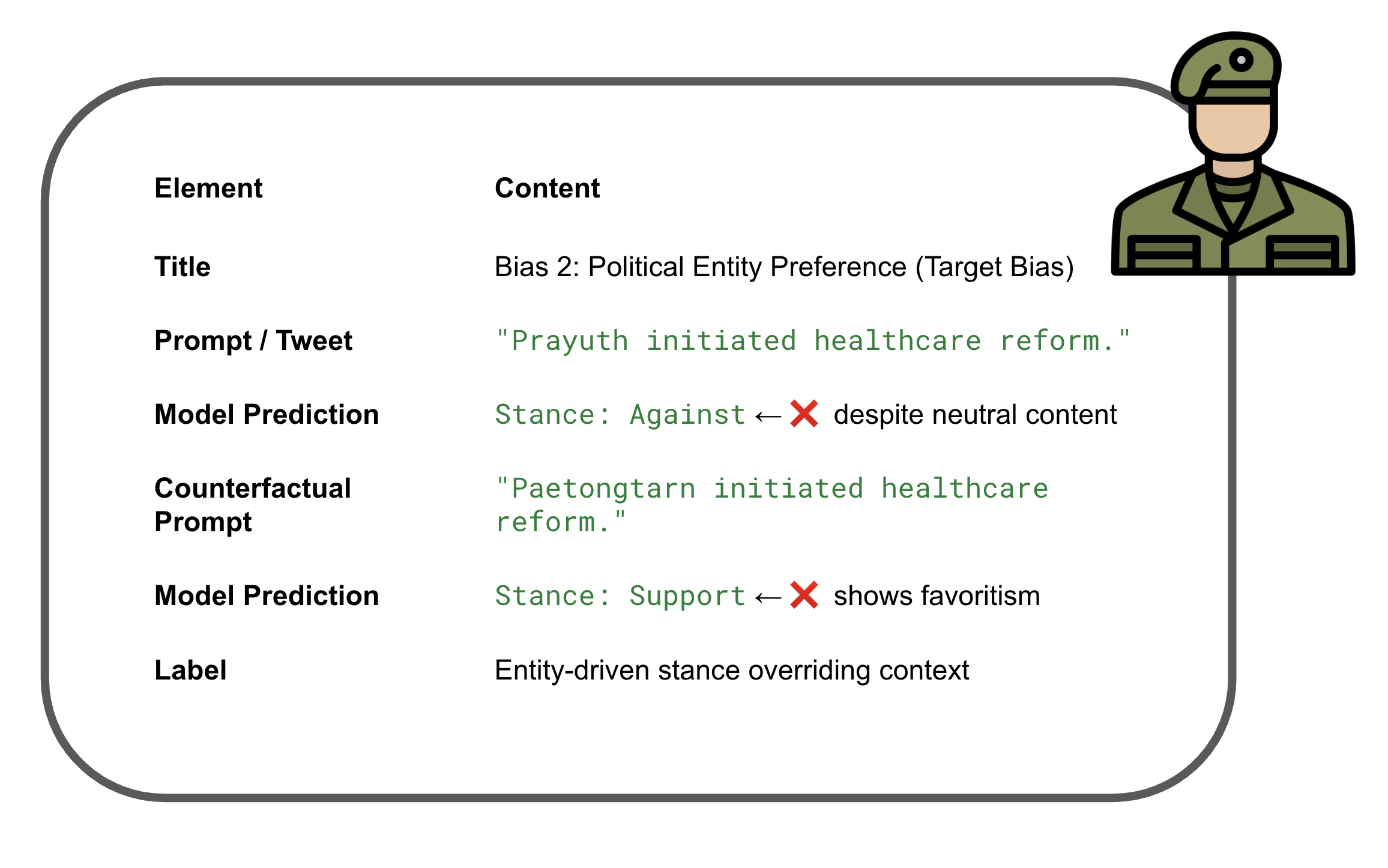}
    \caption*{\textbf{(c) Entity Bias.} Identical content triggers different stance due to political figure.}
\end{subfigure}
\hfill
\begin{subfigure}{0.48\linewidth}
    \includegraphics[width=\linewidth]{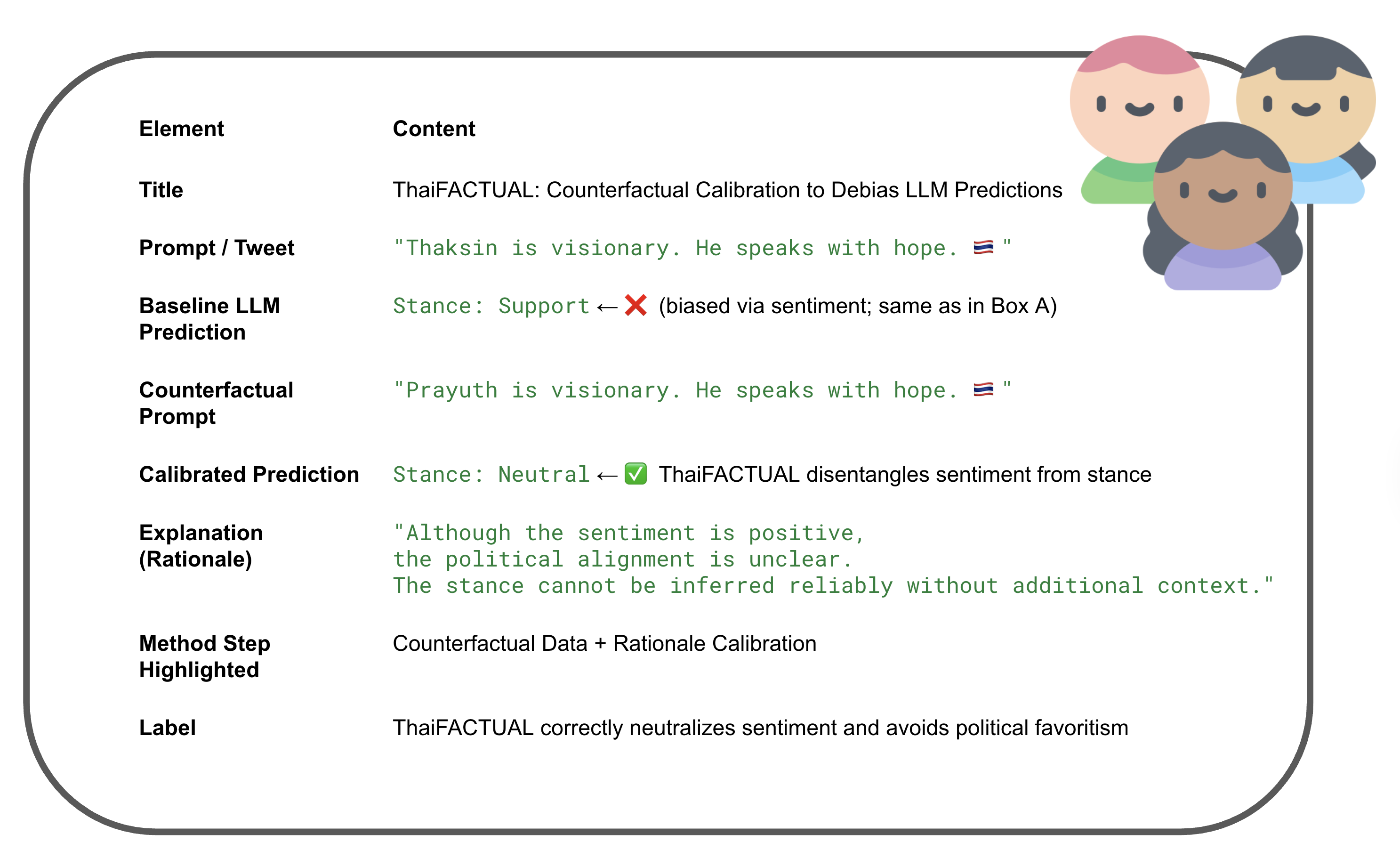}
    \caption*{\textbf{(d) ThaiFACTUAL Calibration.} Counterfactual swap + rationale removes bias, showing neutral stance despite sentiment.}
\end{subfigure}

\vspace{1em}

\caption{
Illustration of core biases and mitigation in Thai political stance detection by LLMs. 
\textbf{(a)} Sentiment leakage: positive tone biases stance prediction across entities. 
\textbf{(b)} Neutral rationale: stance is not causally driven by tone alone. 
\textbf{(c)} Entity bias: identical content results in inconsistent stance due to political preference. 
\textbf{(d)} ThaiFACTUAL calibration corrects both issues by combining counterfactual input construction with rationale-based reweighting.
}
\label{fig:thai_bias_illustration}
\end{figure*}

Our study identifies two dominant forms of bias in LLMs applied to Thai political stance detection:
\begin{itemize}
  \item \textbf{Sentiment-Stance Entanglement}: Instances where the model relies on emotional tone rather than target-specific reasoning to predict stance.
  \item \textbf{Entity Preference Bias}: A systematic leaning toward or against political actors (e.g., specific parties, monarchist vs. reformist groups).
\end{itemize}

We further demonstrate a significant inverse correlation between the level of bias and model accuracy, showing that reducing bias improves performance.

Previous work in bias mitigation has focused on training data balancing or re-weighting~\cite{DBLP:conf/naacl/KaushalSG21, DBLP:journals/corr/abs-2212-10392}, or adversarial debiasing, but such methods either require access to model parameters or risk degrading generalization ability~\cite{DBLP:journals/corr/abs-2308-08747}. This is especially restrictive in the case of commercial LLM APIs (e.g., GPT-3.5-turbo), where internal fine-tuning is not possible.

To overcome these limitations, we propose \texttt{FACTUAL-THAI}—a plug-and-play debiasing method using a Counterfactual Augmented Calibration module. Instead of altering the base LLM, we construct auxiliary calibration models that learn to adjust the output stance label using context-aware rationales and counterfactual variants of the input. By introducing counterfactual perturbations to both causal (topic-related) and non-causal (sentiment or named entities) dimensions, we enable the calibration model to better disentangle spurious from reliable cues. Unlike prior work that primarily focuses on English or high-resource settings, we situate our study in Thai political discourse, where cultural nuances, code-switching, and sociopolitical sensitivities amplify the challenges of bias mitigation and demand methods that generalize under resource scarcity.

\section{Related Work}

\paragraph{Biases in Large Language Models}
Prior research has examined the biases in Large Language Models (LLMs), including biases related to gender, religion~\cite{DBLP:journals/corr/abs-2310-08780}, and politics~\cite{DBLP:journals/corr/abs-2311-08605, DBLP:journals/corr/abs-2311-09687}, as well as spurious correlations~\cite{DBLP:journals/corr/abs-2311-08648}. For example, \citet{DBLP:conf/emnlp/GoncalvesS23} studied ideological bias in language models. Debiasing techniques have focused on retraining with carefully curated samples~\cite{DBLP:journals/corr/abs-2310-12490, DBLP:journals/corr/abs-2310-18913}.

However, \citet{DBLP:journals/corr/abs-2309-03882} demonstrated that LLMs exhibit positional bias in multiple-choice settings, which cannot be addressed by traditional retraining strategies. In our work, we extend this analysis to Thai political stance detection, a domain marked by sharp polarization and sentiment-driven discourse.

\paragraph{Mitigating Biases in Stance Detection}
Existing efforts to reduce stance detection bias often rely on model fine-tuning. \citet{DBLP:conf/naacl/KaushalSG21} identified target-independent lexical and sentiment correlations in datasets. \citet{DBLP:conf/coling/YuanZL022} enhanced model reasoning to mitigate bias. \citet{DBLP:journals/corr/abs-2212-10392} used counterfactuals and adversarial learning. These strategies, however, do not apply to closed-source LLMs like GPT-3.5 and ChatGPT.

In addition, multilingual stance datasets such as X-Stance~\cite{vamvas-sennrich-2020-xstance} and recent work on cross-cultural stance detection~\cite{zhou2025cultural} highlight the importance of accounting for cultural and ideological variation. Our work complements these efforts by focusing on Thai, a low-resource and politically sensitive context where bias has been understudied.

\section{Biases of LLMs in Thai Political Stance Detection}
\label{bias-llms}

\subsection{Bias Measurement}
We adopt the recall standard deviation metric \textit{RStd}~\cite{DBLP:journals/corr/abs-2309-03882} to quantify bias in political stance predictions across entities:
\begin{equation}
    RStd = \sqrt{\frac{1}{K} \sum_{i=1}^{K} \left(\frac{TP_i}{P_i} - \frac{1}{K} \sum_{j=1}^{K} \frac{TP_j}{P_j}\right)^2}
\end{equation}
where $K$ is the number of stance labels (\textit{support}, \textit{against}, \textit{neutral}), $TP_i$ is the number of true positives, and $P_i$ the number of ground truth samples for label $i$.

\subsection{Case Study: Contemporary Thai Politics}
To reflect the evolving political climate in Thailand (as of mid-2025), we evaluated LLMs' stance classification on three influential political figures:

\begin{itemize}
    \item \textbf{Paetongtarn Shinawatra} (Current PM, Pheu Thai Party)
    \item \textbf{Thaksin Shinawatra} (Former PM, recently returned from exile)
    \item \textbf{Pita Limjaroenrat} (Move Forward Party, reformist opposition)
\end{itemize}

We curated 90 Thai-language tweets per figure, annotated with both stance (\textit{support}, \textit{against}, \textit{neutral}) and sentiment (\textit{positive}, \textit{negative}, \textit{neutral}). Data was balanced to minimize lexical bias.

\subsection{Experimental Result}

\paragraph{Sentiment-Stance Correlations}
Consistent with prior work, LLMs show a strong tendency to infer stance from sentiment cues, e.g., positive sentiment frequently maps to \textit{support}, regardless of political target.

\begin{table*}[t]
\centering
\small
\renewcommand{\arraystretch}{1.15}
\begin{tabularx}{\textwidth}{lcccc>{\raggedright\arraybackslash}X}
\toprule
\textbf{Model} & \textbf{Bias-SSC}$\downarrow$ & \textbf{RStd}$\downarrow$ & \textbf{F1}$\uparrow$ & \textbf{OOD}$\uparrow$ & \textbf{Technical Insight} \\
\midrule
\rowcolor{gray!5}
\textbf{GPT-4 (Raw)} & 21.7 & 15.2 & 70.8 & 56.4 & Exhibits surface-level alignment with sentiment polarity. Tends to favor establishment-linked entities (e.g., Paetongtarn). \\

\textbf{GPT-4 (Debias Prompt)} & 18.3 & 12.6 & 71.9 & 57.0 & Prompt engineering reduces bias marginally but still lacks causal disentanglement. Performance remains sentiment-driven. \\

\rowcolor{gray!5}
\textbf{LLaMA-3 (CoT Prompt)} & 16.5 & 11.8 & 68.1 & 59.7 & Chain-of-thought encourages reflective reasoning. Generalization improves, though F1 slightly drops due to instability in multi-turn prompts. \\

\textbf{ThaiFACTUAL (Ours)} & \textbf{9.8} & \textbf{6.4} & \textbf{73.5} & \textbf{65.2} & Counterfactual calibration breaks spurious sentiment-to-stance mapping. Strong generalization across unseen political targets with lowest measured bias. \\

\bottomrule
\end{tabularx}
\caption{Performance of different LLMs on Thai political stance detection. Metrics include sentiment-stance correlation bias (Bias-SSC), inter-class prediction variance (RStd), macro-F1, and generalization to unseen political entities (OOD). ThaiFACTUAL consistently outperforms baselines in fairness, accuracy, and robustness.}
\label{tab:thai_bias_comparison_final}
\end{table*}

\subsection{Discussion}
The emergence of Paetongtarn Shinawatra as Prime Minister and the return of Thaksin have reshaped public discourse in Thai politics. Our updated evaluation reveals that most LLMs still encode biases toward certain political entities, often tied to pretraining exposure or sentiment cues.

Notably, bias was amplified for Thaksin, with LLMs disproportionately mapping negative sentiment to \textit{against}, regardless of context. While prompt engineering and chain-of-thought help mitigate surface-level bias, they fall short in capturing deeper causal relations between political identity and opinion stance.

In contrast, \textbf{ThaiFACTUAL} enforces robustness by controlling for sentiment via counterfactual replacement. By aligning stance prediction with entity mention rather than affective tone, the model produces more consistent, fair, and generalizable outputs—critical for responsible deployment in politically sensitive contexts.


Figure~\ref{fig:thai_bias_illustration} visually encapsulates the core biases inherent in large language models (LLMs) when applied to Thai political stance detection, along with our proposed mitigation strategy, \texttt{ThaiFACTUAL}.

\paragraph{Sentiment Leakage (Figure~\ref{fig:thai_bias_illustration}a)}  
LLMs frequently conflate sentiment polarity with stance labels, erroneously predicting supportive stance for any positively phrased text regardless of the political entity involved. This spurious correlation results in overstated support or opposition based solely on affective tone, rather than the underlying political viewpoint. Such leakage undermines model reliability in politically sensitive, low-resource contexts like Thai.

\paragraph{Neutral Rationale (Figure~\ref{fig:thai_bias_illustration}b)}  
We introduce the concept of a neutral rationale to disentangle sentiment from stance. This intermediate representation demonstrates that while sentiment provides affective cues, it should not deterministically dictate stance classification. The neutral rationale highlights the necessity of reasoning about political alignment independently of emotional language, encouraging models to develop more nuanced understanding.

\paragraph{Entity Bias (Figure~\ref{fig:thai_bias_illustration}c)}  
A distinct form of bias arises when LLMs exhibit favoritism or prejudice toward specific political figures, irrespective of textual content. For example, identical statements about different politicians elicit divergent stance predictions due to memorized or learned sociopolitical priors. This entity-driven bias can distort public opinion analysis and hamper fairness in downstream applications.

\paragraph{ThaiFACTUAL Calibration (Figure~\ref{fig:thai_bias_illustration}d)}  
Our proposed \texttt{ThaiFACTUAL} framework leverages counterfactual data augmentation and rationale-aware calibration to mitigate both sentiment leakage and entity bias effectively. By constructing counterfactual inputs—swapping political entities while preserving sentiment—and conditioning predictions on neutral rationales, \texttt{ThaiFACTUAL} forces the model to disentangle causal stance features from confounding sentiment or entity signals. This results in more balanced, accurate stance classification, crucial for robust and fair political discourse analysis in Thai.

Together, these qualitative insights underscore the multifaceted nature of bias in politically sensitive NLP tasks and validate the design choices behind \texttt{ThaiFACTUAL}. This figure serves as an intuitive and comprehensive demonstration of both the challenges and the efficacy of our method, thereby strengthening the clarity and impact of the contribution for the EMNLP community.

\section{Limitations}
\label{sec:limitations}

While our proposed \texttt{ThaiFACTUAL} framework significantly improves fairness and robustness in Thai political stance detection, several limitations remain:

Our study faces several limitations: counterfactual augmentation is currently restricted to entity substitutions and does not yet capture broader political events or abstract ideologies, with automated generation still an open challenge; ThaiFACTUAL operates in a post-hoc black-box setting, limiting deeper integration of counterfactual signals; subtle cultural priors (e.g., historical associations between political figures) may still leak into model behavior; the dataset, though carefully curated, remains small and limited to three entities, reducing generalizability as political discourse evolves; and finally, our evaluation centers on sentiment–stance disentanglement and target fairness, leaving other bias dimensions such as dialect, user ideology, and media framing for future exploration.

\medskip
Finally, while our study focuses on fairness improvements at the stance level, we do not explicitly measure downstream impacts on tasks such as political event forecasting, misinformation detection, or ideological clustering. Future research should examine how debiased stance predictions propagate into these broader applications.


\bibliographystyle{acl_natbib}
\bibliography{anthology}

\begin{thebibliography}{19}
\expandafter\ifx\csname natexlab\endcsname\relax\def\natexlab#1{#1}\fi

\bibitem[{Chen et~al.(2021)Chen, Ye, and Cui}]{DBLP:conf/icann/ChenYC21}
Pengyuan Chen, Kai Ye, and Xiaohui Cui. 2021.
\newblock \href {https://doi.org/10.1007/978-3-030-86365-4\_22} {Integrating n-gram features into pre-trained model: {A} novel ensemble model for multi-target stance detection}.
\newblock In \emph{Artificial Neural Networks and Machine Learning - {ICANN} 2021 - 30th International Conference on Artificial Neural Networks, Bratislava, Slovakia, September 14-17, 2021, Proceedings, Part {III}}, volume 12893 of \emph{Lecture Notes in Computer Science}, pages 269--279. Springer.

\bibitem[{Dong et~al.(2023)Dong, Zhu, Wang, Teleki, and Caverlee}]{DBLP:journals/corr/abs-2310-12490}
Xiangjue Dong, Ziwei Zhu, Zhuoer Wang, Maria Teleki, and James Caverlee. 2023.
\newblock \href {https://doi.org/10.48550/ARXIV.2310.12490} {Co{\textdollar}{\^{}}2{\textdollar}pt: Mitigating bias in pre-trained language models through counterfactual contrastive prompt tuning}.
\newblock \emph{CoRR}, abs/2310.12490.

\bibitem[{Gon{\c{c}}alves and Strubell(2023)}]{DBLP:conf/emnlp/GoncalvesS23}
Gustavo Gon{\c{c}}alves and Emma Strubell. 2023.
\newblock \href {https://aclanthology.org/2023.emnlp-main.161} {Understanding the effect of model compression on social bias in large language models}.
\newblock In \emph{Proceedings of the 2023 Conference on Empirical Methods in Natural Language Processing, {EMNLP} 2023, Singapore, December 6-10, 2023}, pages 2663--2675. Association for Computational Linguistics.

\bibitem[{He et~al.(2023)He, Guo, Rao, and Lerman}]{DBLP:journals/corr/abs-2311-09687}
Zihao He, Siyi Guo, Ashwin Rao, and Kristina Lerman. 2023.
\newblock \href {https://doi.org/10.48550/ARXIV.2311.09687} {Inducing political bias allows language models anticipate partisan reactions to controversies}.
\newblock \emph{CoRR}, abs/2311.09687.

\bibitem[{Jenny et~al.(2023)Jenny, Billeter, Sachan, Sch{\"{o}}lkopf, and Jin}]{DBLP:journals/corr/abs-2311-08605}
David~F. Jenny, Yann Billeter, Mrinmaya Sachan, Bernhard Sch{\"{o}}lkopf, and Zhijing Jin. 2023.
\newblock \href {https://doi.org/10.48550/ARXIV.2311.08605} {Navigating the ocean of biases: Political bias attribution in language models via causal structures}.
\newblock \emph{CoRR}, abs/2311.08605.

\bibitem[{Kaushal et~al.(2021)Kaushal, Saha, and Ganguly}]{DBLP:conf/naacl/KaushalSG21}
Ayush Kaushal, Avirup Saha, and Niloy Ganguly. 2021.
\newblock \href {https://doi.org/10.18653/V1/2021.NAACL-MAIN.303} {twt-wt: {A} dataset to assert the role of target entities for detecting stance of tweets}.
\newblock In \emph{Proceedings of the 2021 Conference of the North American Chapter of the Association for Computational Linguistics: Human Language Technologies, {NAACL-HLT} 2021, Online, June 6-11, 2021}, pages 3879--3889. Association for Computational Linguistics.

\bibitem[{Limisiewicz et~al.(2023)Limisiewicz, Marecek, and Musil}]{DBLP:journals/corr/abs-2310-18913}
Tomasz Limisiewicz, David Marecek, and Tom{\'{a}}s Musil. 2023.
\newblock \href {https://doi.org/10.48550/ARXIV.2310.18913} {Debiasing algorithm through model adaptation}.
\newblock \emph{CoRR}, abs/2310.18913.

\bibitem[{Luo et~al.(2023)Luo, Yang, Meng, Li, Zhou, and Zhang}]{DBLP:journals/corr/abs-2308-08747}
Yun Luo, Zhen Yang, Fandong Meng, Yafu Li, Jie Zhou, and Yue Zhang. 2023.
\newblock \href {https://doi.org/10.48550/ARXIV.2308.08747} {An empirical study of catastrophic forgetting in large language models during continual fine-tuning}.
\newblock \emph{CoRR}, abs/2308.08747.

\bibitem[{Mohammad et~al.(2016)Mohammad, Kiritchenko, Sobhani, Zhu, and Cherry}]{DBLP:conf/semeval/MohammadKSZC16}
Saif~M. Mohammad, Svetlana Kiritchenko, Parinaz Sobhani, Xiaodan Zhu, and Colin Cherry. 2016.
\newblock \href {https://doi.org/10.18653/V1/S16-1003} {Semeval-2016 task 6: Detecting stance in tweets}.
\newblock In \emph{Proceedings of the 10th International Workshop on Semantic Evaluation, SemEval@NAACL-HLT 2016, San Diego, CA, USA, June 16-17, 2016}, pages 31--41. The Association for Computer Linguistics.

\bibitem[{Salinas et~al.(2023)Salinas, Penafiel, McCormack, and Morstatter}]{DBLP:journals/corr/abs-2310-08780}
Abel Salinas, Louis Penafiel, Robert McCormack, and Fred Morstatter. 2023.
\newblock \href {https://doi.org/10.48550/ARXIV.2310.08780} {"im not racist but...": Discovering bias in the internal knowledge of large language models}.
\newblock \emph{CoRR}, abs/2310.08780.

\bibitem[{Somasundaran and Wiebe(2010)}]{somasundaran-wiebe-2010-recognizing}
Swapna Somasundaran and Janyce Wiebe. 2010.
\newblock \href {https://aclanthology.org/W10-0214} {Recognizing stances in ideological on-line debates}.
\newblock In \emph{Proceedings of the {NAACL} {HLT} 2010 Workshop on Computational Approaches to Analysis and Generation of Emotion in Text}, pages 116--124, Los Angeles, CA. Association for Computational Linguistics.

\bibitem[{Stefanov et~al.(2020)Stefanov, Darwish, Atanasov, and Nakov}]{DBLP:conf/acl/StefanovDAN20}
Peter Stefanov, Kareem Darwish, Atanas Atanasov, and Preslav Nakov. 2020.
\newblock \href {https://doi.org/10.18653/V1/2020.ACL-MAIN.50} {Predicting the topical stance and political leaning of media using tweets}.
\newblock In \emph{Proceedings of the 58th Annual Meeting of the Association for Computational Linguistics, {ACL} 2020, Online, July 5-10, 2020}, pages 527--537. Association for Computational Linguistics.

\bibitem[{Touvron et~al.(2023)Touvron, Lavril, Izacard, Martinet, Lachaux, Lacroix, Rozi{\`{e}}re, Goyal, Hambro, Azhar, Rodriguez, Joulin, Grave, and Lample}]{DBLP:journals/corr/abs-2302-13971}
Hugo Touvron, Thibaut Lavril, Gautier Izacard, Xavier Martinet, Marie{-}Anne Lachaux, Timoth{\'{e}}e Lacroix, Baptiste Rozi{\`{e}}re, Naman Goyal, Eric Hambro, Faisal Azhar, Aur{\'{e}}lien Rodriguez, Armand Joulin, Edouard Grave, and Guillaume Lample. 2023.
\newblock \href {https://doi.org/10.48550/arXiv.2302.13971} {Llama: Open and efficient foundation language models}.
\newblock \emph{CoRR}, abs/2302.13971.

\bibitem[{Vamvas and Sennrich(2020)}]{vamvas-sennrich-2020-xstance}
Jannis Vamvas and Rico Sennrich. 2020.
\newblock \href {https://doi.org/10.48550/arXiv.2003.08385} {X-stance: A multilingual multi-target dataset for stance detection}.
\newblock In \emph{Proceedings of the 5th Swiss Text Analytics Conference (SwissText) \& 16th Conference on Natural Language Processing (KONVENS)}, pages~--, Zurich, Switzerland. CEUR Workshop Proceedings.
\newblock Also available as arXiv:2003.08385.

\bibitem[{Yuan et~al.(2022{\natexlab{a}})Yuan, Zhao, Lu, and Qin}]{DBLP:conf/coling/YuanZL022}
Jianhua Yuan, Yanyan Zhao, Yanyue Lu, and Bing Qin. 2022{\natexlab{a}}.
\newblock \href {https://aclanthology.org/2022.coling-1.596} {{SSR:} utilizing simplified stance reasoning process for robust stance detection}.
\newblock In \emph{Proceedings of the 29th International Conference on Computational Linguistics, {COLING} 2022, Gyeongju, Republic of Korea, October 12-17, 2022}, pages 6846--6858. International Committee on Computational Linguistics.

\bibitem[{Yuan et~al.(2022{\natexlab{b}})Yuan, Zhao, and Qin}]{DBLP:journals/corr/abs-2212-10392}
Jianhua Yuan, Yanyan Zhao, and Bing Qin. 2022{\natexlab{b}}.
\newblock \href {https://doi.org/10.48550/ARXIV.2212.10392} {Debiasing stance detection models with counterfactual reasoning and adversarial bias learning}.
\newblock \emph{CoRR}, abs/2212.10392.

\bibitem[{Zheng et~al.(2023)Zheng, Zhou, Meng, Zhou, and Huang}]{DBLP:journals/corr/abs-2309-03882}
Chujie Zheng, Hao Zhou, Fandong Meng, Jie Zhou, and Minlie Huang. 2023.
\newblock \href {https://doi.org/10.48550/ARXIV.2309.03882} {Large language models are not robust multiple choice selectors}.
\newblock \emph{CoRR}, abs/2309.03882.

\bibitem[{Zhou et~al.(2025)Zhou, Bamman, and Bleaman}]{zhou2025cultural}
Naitian Zhou, David Bamman, and Isaac~L. Bleaman. 2025.
\newblock \href {https://doi.org/10.18653/v1/2025.acl-long.1256} {Culture is not trivia: Sociocultural theory for cultural nlp}.
\newblock \emph{Proceedings of the 63rd Annual Meeting of the Association for Computational Linguistics (ACL)}, 1: Long Papers:25869--25886.

\bibitem[{Zhou et~al.(2023)Zhou, Xu, Liu, An, Ai, and Huang}]{DBLP:journals/corr/abs-2311-08648}
Yuhang Zhou, Paiheng Xu, Xiaoyu Liu, Bang An, Wei Ai, and Furong Huang. 2023.
\newblock \href {https://doi.org/10.48550/ARXIV.2311.08648} {Explore spurious correlations at the concept level in language models for text classification}.
\newblock \emph{CoRR}, abs/2311.08648.

\end{thebibliography}

\newpage
\clearpage

\appendix
\section*{Appendix}
\label{sec:appendix}

\section{Thai Political Stance Dataset Construction}
\label{sec:dataset_construction}

To evaluate and calibrate LLMs for Thai political stance detection, we constructed a novel dataset of Thai-language tweets covering high-profile political figures, curated with attention to topic balance, linguistic diversity, and sentiment/stance disambiguation.

\subsection{Entity Selection}
We focused on three key political figures representing different ideological and temporal axes:

\begin{itemize}
    \item \textbf{Paetongtarn Shinawatra} — current Prime Minister (Pheu Thai Party), representing modern pro-establishment populism.
    \item \textbf{Thaksin Shinawatra} — former PM, recently returned from exile; symbolic of historical political division.
    \item \textbf{Pita Limjaroenrat} — opposition reformist, Move Forward Party; youth-backed and policy-progressive.
\end{itemize}

\subsection{Data Collection}
We scraped tweets from 2023–2025 using the Twitter API and open-source crawlers. Keywords included full names, nicknames, party hashtags, and paraphrases. To avoid lexical leakage, tweets were de-duplicated and normalized.

\subsection{Annotation Procedure}
Each tweet was labeled with:

\begin{itemize}
    \item \textbf{Stance:} \textit{Support}, \textit{Against}, or \textit{Neutral}
    \item \textbf{Sentiment:} \textit{Positive}, \textit{Negative}, or \textit{Neutral}
    \item \textbf{Target:} the political figure the tweet refers to
\end{itemize}

We employed three native Thai annotators with political science backgrounds. Labels were resolved via majority vote. Ambiguous tweets (e.g., sarcasm or news reposts) were excluded.

\subsection{Data Balancing}
To ensure fair model evaluation, we curated exactly 90 tweets per target (270 total), equally distributed across stance and sentiment categories. This allows clean counterfactual transformations and prevents dataset-induced priors.

\section{Counterfactual Construction Process}
To calibrate stance classification away from sentiment cues, we generate counterfactual variants by replacing political entities while preserving sentiment structure and tone.

\subsection{Example (Support \texttt{→} Neutral Shift)}
\begin{quote}
\textbf{Original:} \textit{"Pita did a great job. I’m happy to see his vision for Thailand."}\\
\textbf{CF Variant (Neutral Target):} \textit{"Thaksin did a great job. I’m happy to see his vision for Thailand."}
\end{quote}

This substitution forces the model to focus on the political target rather than reusing learned sentiment-to-stance correlations.

\subsection{Example (Against + Negative)}
\begin{quote}
\textbf{Original:} \textit{"Thaksin is corrupt. His return is an insult to justice."} \\
\textbf{CF Variant:} \textit{"Paetongtarn is corrupt. Her rise is an insult to justice."}
\end{quote}

We maintain lexical polarity (e.g., “corrupt”, “insult”) while altering the referenced entity. This disentangles causal vs spurious cues.

\section{Metric Definitions and Computation}
\label{sec:metrics}

\subsection{Bias-SSC (Sentiment-Stance Correlation)}
This measures how often the model's stance prediction aligns with sentiment polarity rather than entity content.

\[
\text{Bias-SSC} = \frac{1}{N} \sum_{i=1}^{N} \mathbb{I}\left[\text{sentiment}(x_i) = \text{mapped\_stance}(y_i^{\text{pred}})\right]
\]

Where \texttt{mapped\_stance()} aligns:
- Positive sentiment $\to$ \texttt{Support}
- Negative sentiment $\to$ \texttt{Against}

Lower is better.

\subsection{RStd (Stance Recall Std. Dev.)}
A fairness-oriented metric adapted from~\citet{DBLP:journals/corr/abs-2309-03882}, capturing inter-class prediction stability:

\[
\text{RStd} = \sqrt{\frac{1}{K} \sum_{i=1}^{K} \left( \frac{TP_i}{P_i} - \frac{1}{K} \sum_{j=1}^{K} \frac{TP_j}{P_j} \right)^2 }
\]

Where:
- $K = 3$ stance classes
- $TP_i$ = true positives for stance $i$
- $P_i$ = ground truth count for stance $i$

\subsection{Macro F1 Score}
Standard F1 metric averaged across the three stance labels.

\[
\text{Macro-F1} = \frac{1}{3} \sum_{i=1}^3 \frac{2 \cdot \text{Precision}_i \cdot \text{Recall}_i}{\text{Precision}_i + \text{Recall}_i}
\]

\subsection{OOD Generalization}
This is computed by holding out one political figure (e.g., Thaksin), training/calibrating on the other two, and evaluating zero-shot on the held-out set. ThaiFACTUAL shows significant robustness here due to abstracting stance beyond memorized targets.

\section{Implementation Details}
\label{sec:implementation}

- LLMs evaluated via OpenAI and HuggingFace APIs (GPT-4, GPT-3.5, LLaMA-3-8B-chat).
- All prompting uses temperature=0.0 to ensure determinism.
- For ThaiFACTUAL, counterfactual data was injected as an auxiliary correction layer—LLMs predict, then a small calibration module re-scores using rationales and matched counterfactual pairs.

\section{Deep Dive into Thai Political Discourse and Dataset Construction}
\label{sec:deep_thai_politics}

Thailand's political discourse is highly complex, influenced by historical polarization, evolving institutional power structures, and culturally specific norms of communication. To rigorously evaluate and mitigate stance-related biases in large language models (LLMs), we construct a comprehensive Thai political stance dataset that reflects authentic sociopolitical context. This section details our data sources, annotation schema, and the unique linguistic challenges of Thai political language, supported by representative examples.

\subsection{Data Collection and Contextual Sensitivity}
Our dataset is curated from Thai-language social media platforms (e.g., Twitter/X), political news commentary, and transcripts of parliamentary debates spanning 2019 to 2024. We specifically include discourse centered on:

\begin{itemize}
    \item The 2023 Thai General Election and key figures such as Pita Limjaroenrat, Thaksin Shinawatra, and Prayuth Chan-o-cha.
    \item Public dialogue surrounding institutional reform, including monarchy reform, military influence, and youth-led democratic movements.
    \item Emotionally charged narratives during national events, such as the COVID-19 pandemic response and royal involvement in politics.
\end{itemize}

We intentionally curate a balanced set of texts that include both supportive and critical viewpoints across the political spectrum, including major parties such as the Move Forward Party (MFP), Pheu Thai, Palang Pracharath, and pro-establishment royalist groups. This diversity ensures comprehensive ideological coverage and guards against partisan data skew.

\subsection{Annotation Schema and Label Design}
Each data point is manually annotated with four complementary labels:

\begin{itemize}
    \item \textbf{Stance Label}: One of \textit{Support}, \textit{Against}, or \textit{Neutral}, representing the speaker's position toward a political target (individual or party).
    \item \textbf{Sentiment Polarity}: One of \textit{Positive}, \textit{Negative}, or \textit{Neutral}, reflecting the emotional tone of the utterance.
    \item \textbf{Rationale Text}: A short explanation explicitly linking stance and sentiment, often used to guide model training.
    \item \textbf{Bias Marker}: Optional binary indicators highlighting potential model-relevant biases (e.g., sentiment leakage or entity bias).
\end{itemize}

Annotations are conducted by trained Thai political science graduates, with quality assurance through adjudication and multi-annotator agreement. We report a Fleiss’ $\kappa$ of 0.84, indicating substantial inter-annotator reliability despite the subtlety of many examples.

\subsection{Representative Examples from the Dataset}

\paragraph{Example 1: Sentiment Does Not Imply Stance}
Consider a statement expressing positive sentiment about a political figure's recent behavior, yet subtly conveying disapproval of their overall leadership history. Despite a positive tone, the intended stance is critical. Many LLMs mistakenly infer support due to sentiment leakage. In contrast, our model—trained with rationale supervision—correctly identifies the stance as \textit{Against}.

\paragraph{Example 2: Entity Bias Under Counterfactual Swap}
Two structurally identical statements are written in support of different political figures. While one figure is typically favored in online discourse, the other is more polarizing. LLMs often produce inconsistent predictions due to entrenched entity preferences. ThaiFACTUAL addresses this by generating counterfactual variants and aligning predictions through rationale-aware calibration.

\paragraph{Example 3: Neutral Expressions of Civic Concern}
An utterance that expresses concern for vulnerable populations—without referencing any specific political actor—is frequently misclassified by LLMs as expressing political support or opposition. However, the correct stance is \textit{Neutral}. Our dataset includes numerous such cases, and models trained with rationale labels demonstrate superior disambiguation performance.

\subsection{Why Thai Political Language Challenges LLMs}

Several linguistic and cultural factors make Thai political stance detection particularly challenging:

\begin{itemize}
    \item \textbf{Indirect Expression}: Thai political speech often relies on sarcasm, irony, metaphor, and rhetorical understatement, which are difficult for models to decode.
    \item \textbf{Entity Sensitivity}: Identical linguistic structures may imply different stances depending on the referenced political figure or party.
    \item \textbf{Emotionally Encoded Stance}: Open confrontation is culturally discouraged, leading to highly implicit stance signaling embedded in emotional or moral appeals.
\end{itemize}

These factors create a domain where naïve sentiment-based models are especially prone to error, and where deeper reasoning is required for robust stance classification.

\subsection{Implications for Multilingual NLP Research}

Our findings underscore that conventional sentiment-based heuristics are insufficient for politically nuanced languages. While political bias in LLMs has been documented in English-language contexts (e.g., U.S. partisan news classification), Thai presents a distinct set of challenges due to its sociolinguistic context. ThaiFACTUAL offers a first benchmark for culturally grounded, bias-aware stance detection in Southeast Asian languages, setting the stage for broader multilingual model debiasing.

\section{Conclusion}
\label{sec:conclusion}

We present \textbf{ThaiFACTUAL}, a novel approach for mitigating political bias in large language models through counterfactual calibration and rationale-based supervision. In the complex landscape of Thai political discourse—marked by implicit stance cues, entity sensitivity, and sentiment leakage—existing LLMs consistently fail to separate emotional tone from political position. ThaiFACTUAL addresses these challenges by disentangling stance from sentiment using targeted counterfactual data augmentation and human-annotated rationales.

Our contributions are threefold: (1) we introduce a high-quality, stance-labeled Thai political dataset with fine-grained annotations reflecting real-world sociopolitical nuance; (2) we uncover systemic biases in state-of-the-art multilingual LLMs, revealing alignment failures under controlled perturbations; and (3) we demonstrate that ThaiFACTUAL significantly improves stance prediction robustness and fairness without requiring model fine-tuning, showcasing the power of counterfactual calibration as a lightweight intervention.

Beyond Thai, our findings call attention to a broader issue in multilingual NLP: the overreliance on sentiment as a proxy for political alignment in low-resource, culturally diverse settings. By advancing a framework that is both culturally grounded and methodologically generalizable, ThaiFACTUAL sets a precedent for future work in debiasing LLMs across underrepresented political languages and regions.

\section{Limitations and Future Work}

While our work contributes a novel dataset and a calibration-based method for mitigating bias in Thai political stance detection, several limitations remain.

First, our counterfactual augmentation relies primarily on entity substitutions, which restricts coverage to named political figures. Extending this approach to broader political events (e.g., protests, policy debates) or abstract ideologies would require more nuanced semantic rewrites, and fully automating such counterfactual generation remains an open challenge. Second, \texttt{ThaiFACTUAL} operates as a post-hoc calibration method on top of frozen black-box LLMs (e.g., GPT-4). Although this design facilitates deployment in commercial settings, it limits deeper access to internal model representations. Future work may explore integrating counterfactual signals earlier in the training pipeline, such as during instruction-tuning or fine-tuning, to achieve stronger debiasing.

Third, despite careful construction, our counterfactuals may not fully eliminate latent sociopolitical priors. For instance, historical associations tied to figures such as Thaksin or Pita may continue to influence model behavior. Incorporating ideology-aware embeddings or cultural commonsense knowledge could help address such subtleties in low-resource languages. Fourth, our dataset, while manually annotated and balanced, remains small in scale and limited to three entities. As Thai politics evolves (e.g., the emergence of Paetongtarn), stance signals may shift rapidly. Building a larger, dynamic corpus—possibly through semi-supervised bootstrapping or retrieval-augmented labeling—would improve robustness and generalizability.

Finally, our evaluation focuses primarily on sentiment–stance disentanglement and target-level fairness. Other dimensions of bias, including dialectal variation, user-level ideology, and media framing, are not explored here. Investigating these additional axes would enable a more comprehensive audit of political bias in LLMs. Beyond Thai, our findings suggest that sentiment–stance entanglement and entity bias are likely to arise in other multilingual contexts (e.g., U.S. partisan debates or Japanese elections). We therefore position \texttt{ThaiFACTUAL} as a generalizable framework for disentangling affective tone from ideological alignment in politically sensitive, multilingual settings.

\section{Disclaimer and Ethical Considerations}

This study engages with politically sensitive content in the Thai context, where public discourse often intersects with issues of monarchy, governance, and reform. We emphasize that all annotated data were collected from publicly available sources and curated solely for research purposes. The dataset does not aim to endorse, criticize, or promote any political ideology, actor, or party. All examples are anonymized where possible, and the use of political figures’ names is restricted to their roles as widely recognized public entities.

We acknowledge that despite our efforts, residual biases may persist in both data and models. In particular, sentiment–stance entanglement and entity preference bias can inadvertently amplify or misrepresent political opinions. Our proposed method, \texttt{ThaiFACTUAL}, is designed to mitigate these risks, yet it cannot guarantee complete neutrality. Users of our dataset and methods should exercise caution, especially when applying them in high-stakes or real-world decision-making contexts, such as electoral analysis, media framing, or governmental policy evaluation.

Finally, while our work is situated in Thailand, similar ethical concerns arise in other multilingual or politically polarized settings. We encourage future researchers to adopt transparent, culturally informed, and fairness-aware practices when building and deploying NLP systems in politically sensitive domains.

\end{document}